\documentclass[11pt,a4paper]{article}
\usepackage[hyperref]{acl2020}
\usepackage{times}
\usepackage{latexsym}

\usepackage{url}

\usepackage{tikz-dependency}
\usepackage{enumitem}
\usepackage{bm}
\usepackage{amsmath}
\usepackage{amssymb}
\usepackage{graphicx}
\graphicspath{ {./plots/} }
\usepackage{color}
\usepackage{tabularx}
\usepackage{booktabs}
\usepackage{multirow}
\usepackage{xfrac}
\usepackage[toc,page]{appendix}

\usepackage{subcaption}
\usepackage[labelsep=quad,skip=3pt]{caption}

\usepackage{microtype}

\aclfinalcopy 
\def\aclpaperid{35} 

\definecolor{DarkRed}{RGB}{140,5,0}

\newcommand{\ie}{i.\,e.\ }
\newcommand{\eg}{e.\,g.\ }

\newcommand{\vs}{vs.\ }

\title{The ADAPT Enhanced Dependency Parser at the 
IWPT 2020 Shared Task}

\author{James Barry \and Joachim Wagner \and Jennifer Foster \\
  ADAPT Centre \\ 
  School of Computing,
  Dublin City University, Ireland \\
  {\tt firstname.lastname@adaptcentre.ie}
}

\date{}

\begin{document}
\maketitle
\begin{abstract}

We describe the ADAPT system
for the 2020 IWPT Shared Task on
parsing enhanced Universal Dependencies
in 17 languages.
We implement a pipeline approach
using UDPipe and UDPipe-future
to provide initial levels of annotation.
The enhanced dependency graph is either produced by
a graph-based semantic dependency parser
or is built from the basic tree using a small set of heuristics.
Our results show that, for the majority of languages,
a semantic dependency parser can be successfully applied to the task of
parsing enhanced dependencies.

Unfortunately, we did not ensure a connected graph as part of our pipeline approach and
our competition submission relied on a last-minute fix
to pass the validation script
which harmed our official evaluation scores significantly.
Our submission ranked eighth in the official evaluation with a macro-averaged coarse ELAS F1 of 67.23 and a treebank average of 67.49. 
We later implemented our own graph-connecting fix  which resulted in a score of 79.53
(language average) or 79.76 (treebank average),
which would have placed fourth in the competition evaluation.

\end{abstract}

\section{Introduction}
\label{sec:intro}

The 
2020 IWPT Shared Task on enhanced dependency parsing
\cite{bouma-etal-2020-overview}
requires participants
to predict the enhanced dependencies
(DEPS column in the CoNLL-U format)
in addition to sentence boundaries, tokenisation, lemmata, POS tags,
morphological features and the basic dependency tree.
We take a pipeline approach using
\begin{enumerate}
    \item UDPipe for sentence splitting and tokenisation,
    \item ensembles of UDPipe-future basic parsers, that also
          predict lemmata, POS tags and morphological features,
          with added support for multi-treebank models
          \cite{stymne-etal-2018-parser}, and
    \item two types of enhancers: \textit{(a)} copying the basic tree and applying a small set of heuristics 
    (baseline system), and \textit{(b)} a graph-based semantic dependency parser \citep{dozat-manning-2018-simpler}.
\end{enumerate}

To enable reproduction of our results, we make available our helper scripts
and modifications of the semantic
parser.\footnote{\url{https://github.com/jbrry/Enhanced-UD-Parsing}}

Our approach to the task does not guarantee a connected graph -- something that we did not account for.
Thus, on submission day, we did not have an appropriate solution ready to fix our outputs
but
were able to provide a valid submission due to some
functionality that was added to the
\texttt{quick-fix}
tool provided by the
organisers\footnote{\url{https://github.com/UniversalDependencies/tools}}
to alter the enhanced graph.
The solution was designed 
primarily
to make the files 
pass validation
but in doing so,
harms F1-score.
In a post-competition run, we addressed the connected graph issue with an alternative solution which increased our macro-averaged ELAS F1-score from 67.23 to 79.53 and
the treebank average from 67.49 to 79.76.


\section{System Components}

\subsection{Segmentation}
\label{sec:segment}

We use UDPipe\footnote{\url{http://ufal.mff.cuni.cz/udpipe}}
\citep{straka-strakova-2017-tokenizing}
with off-the-shelf UD v2.5 models\footnote{\url{http://hdl.handle.net/11234/1-3131}}
\cite{straka-strakova-2019-universal}
for the languages of the shared task
to split the raw input text into sentences and tokens.
In cases where more than one UDPipe model is available for
a language, we try all models
during development
\footnote{Due to a configuration error, we did not try
    segmentation with UDPipe models trained on
    \texttt{fi\_ftb}, \texttt{lt\_hse} and
    \texttt{sv\_lines} in the official submission.
}
and select for each test language
the best overall pipeline according to ELAS
on the treebank with the biggest development set
for the language.\footnote{For
    Czech, we based our decision on results for \texttt{cs\_cac}
    instead of \texttt{cs\_pdt} as we did not have full results available for \texttt{cs\_pdt}.
}


\subsection{Basic Parsing}
\label{sec:basic}

We choose UDPipe-future \cite{straka-2018-udpipe} for basic parsing 
and joint prediction of lemmata, POS tags and morphological features
so as
to not require a separate tagger.
We extend
UDPipe-future to train multi-treebank models
as introduced by \cite{stymne-etal-2018-parser} with UUParser.\footnote{Multi-treebank
    \label{fnt:mtb}
    models supply each token with the
    source treebank ID as additional input with a separate
    embedding table.
    Like \newcite{stymne-etal-2018-parser}, we use a vector size of 12.
    At test time, a proxy treebank must be chosen when the input sentence does not come from one of the training treebanks or the source is unknown.
}$^,$\footnote{\url{https://github.com/jowagner/UDPipe-Future/tree/multitreebank}}

Inspired by \newcite{straka-etal-2019-evaluating},
we use two types of external word embeddings with UDPipe-future:
ELMo contextualised word embeddings \citep{peters-etal-2018-deep}
and
FastText character-n-gram-based word embeddings
\citep{bojanowski-etal-2017-enriching}.\footnote{The
    FastText word embedding is restricted to a fixed vocabulary of
    one million tokens, not taking advantage of FastText's ability to
    produce new vectors for OOVs.
    UDPipe-future does not fine-tune these word embeddings.
    Instead, the parser trains an additional embedding exclusively for
    training words and a character-based representation.
    The latter two are added and the result is concatenated with the
    two externally provided representations.
    As far as we understand the code, an all-zero vector is used for OOVs,
    \ie words not in the selected one-million-word FastText vocabulary.
}
For 15 of the 17 test languages,
ElmoForManyLangs\footnote{\url{https://github.com/HIT-SCIR/ELMoForManyLangs}} \citep{che-etal-2018-towards}
provides ELMo models.
We train FastText
on the raw text provided by the CoNLL'17 shared task for
the same 15 languages after shuffling sentences.
For the Russian FastText model, we kept getting vectors with large
component values even after
trying a different machine and a different permutation of sentences,
prohibiting effective training of the parser.
We then used a model trained on \sfrac{2}{3} of the Russian data
for which component values and parser LAS were in the expected range.
Furthermore, we train UDPipe-future models using FastText and internal embeddings only.

For Lithuanian and Tamil, we train UDPipe-future without external word embeddings.
The parser still uses an internal word embedding covering all words of
the training treebank(s) and a word representation obtained with a 
bidirectional GRU layer over the input characters.

For each target language, we train \textit{(a)} mono-treebank models for
each training treebank available with surface strings in UD v2.5, preferring the shared-task version when available, and
\textit{(b)} a multi-treebank model for each language using all treebanks for that language for which we also trained mono-treebank models.
We train up to seven models with different initialisation
for each setting to combine them in
ensembles.\footnote{We
    trained 68 types of models.
    We trained seven seeds for 34 of these,
    five seeds for 30 and three seeds for four.
    Ensembles sizes three, five and seven are
    considered, including a combination of $(n+1)/2$
    models of one type and $(n-1)/2$ models of
    another type with $n \in \{3,5,7\}$.
}$^,$\footnote{We
    use our implementation
    \url{https://github.com/jowagner/ud-combination}
    of the linear combiner of
    \newcite{attardi-dellorletta-2009-reverse}.
}
We consider ensembles not just of a single type of model with
different initialisation but also combinations of models trained
on different treebanks (mono-treebank models) or treebank combinations
(multi-treebank models) and in the
plain, FastText and ELMo variants.\footnote{While
    predicting on development data to facilitate model selection,
    we temporarily introduced a bug in our system causing it to use
    the first initialisation seed for all ensemble members only,
    effectively falling back to a single model when only one model
    type is used.
    We fixed this bug before we switched to making test set
    predictions and tried to account for it in the model selection
    but, under time pressure, made some hard to explain ad hoc
    choices,
    \eg we used an ensemble of three models for Czech, two mono-treebank
    models trained on \texttt{cs\_cac} and one multi-treebank
    model, even though we also had test set predictions
    with an ensemble of seven models with the same mixture of model
    types available.
    For details, see the reproducibility notes in our code
    repository.
}
As the number of possible combinations increases exponentially
with the number of models, we prune the candidates giving
preference to models using all or only one treebank and
to models using ELMo.
We then test each ensemble on the development data
(raw input segmented with UDPipe) and pick the
best ensembles based on ELAS after applying our heuristic enhancer
(Section~\ref{sec:heuristic})
to the basic trees.

To pick the proxy treebank
(see description in Footnote~\ref{fnt:mtb})
for multi-treebank parsing, we use
the treebank name in the filename of the raw text during development.
However, for final testing, the treebank identifier is unavailable (and
if it had been available there would have been cases where this
treebank is not one of the training treebanks).
Given time limits, we decided to simply assign each test set,
\ie each test language, the training treebank with the largest amount
of training data as the
proxy treebank.\footnote{For Estonian, French, Dutch and Polish
    (a subset of the languages with PUD treebanks announced in the development pack),
    we randomised on the sentence level which proxy treebank
    is used during multi-treebank parsing.
}


\subsection{Heuristic Enhancement}
\label{sec:heuristic}

We build a baseline system which concentrates on the two enhanced UD phenomena which are very straightforward to implement using simple heuristic rules, namely, co-reference in relative clauses and modifier relations containing case markers.
These rules are applied to the output of the basic parser. We have two versions of the modifier relation rule - one in which the value of the \texttt{case} morphological feature is included in the relation label and one without. We also have a rule which adds the lemma of a conjunction to the enhanced label of its head. 
For each development set, we find the optimal subset of the set of heuristic rules in terms of ELAS among all possible subsets except those combining the two case rules.

This baseline system is clearly suboptimal since it makes no attempt at all to handle those more interesting enhanced UD phenomena which involve the addition or deletion of arcs, i.e. conjunct propagation, ellipsis and control/raising constructions.
Nonetheless it is useful as a baseline to check that the main system
is performing reasonably and is available as a fall back.


\subsection{Semantic Parsing}
\label{sec:semantic}

\subsubsection{Modelling Enhanced Dependencies}
As our main system to predict the enhanced graph,
we follow \citep{dozat-manning-2018-simpler} and treat enhanced dependency parsing as a task similar to semantic dependency parsing.
In semantic dependency parsing, words may have multiple heads.
Thus, \newcite{dozat-manning-2018-simpler} apply their deep biaffine graph-based dependency parser \cite{dozat-manning-2017-deep} to the task of semantic dependency parsing but replace the softmax cross-entropy loss with sigmoid cross-entropy loss for edge prediction.
The above modification changes the modelling objective such that words are no longer competing with one another to be classified as the appropriate head;
rather, the parser chooses whether an edge exists between each possible pair of words independently.
Whether an edge exists between two words is based on a predefined threshold, where a score above this threshold results in an edge being predicted
and, subsequently,
the edge's label.
In our experiments we use an edge prediction threshold of 0.5.
If the parser did not predict an edge for a word, we take the edge with the highest probability.
As we want to select the label with the highest probability for each chosen edge, standard softmax cross-entropy loss is used for label prediction as in \newcite{dozat-manning-2018-simpler}.

\begin{figure*}[htb]
  \begin{subfigure}[t]{.5\textwidth}
  \centering\includegraphics[width=.95\columnwidth]{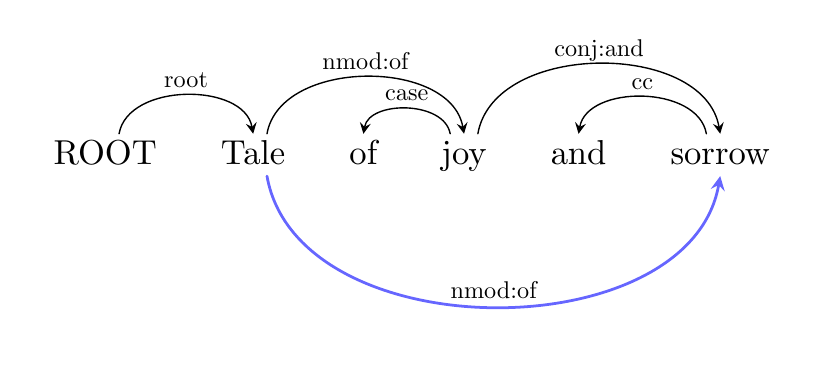}
  \caption{Enhanced UD graph.}
  \label{fig:semantica}
  \end{subfigure}
  \begin{subfigure}[t]{.48\textwidth}
  \centering\includegraphics[width=.95\columnwidth]{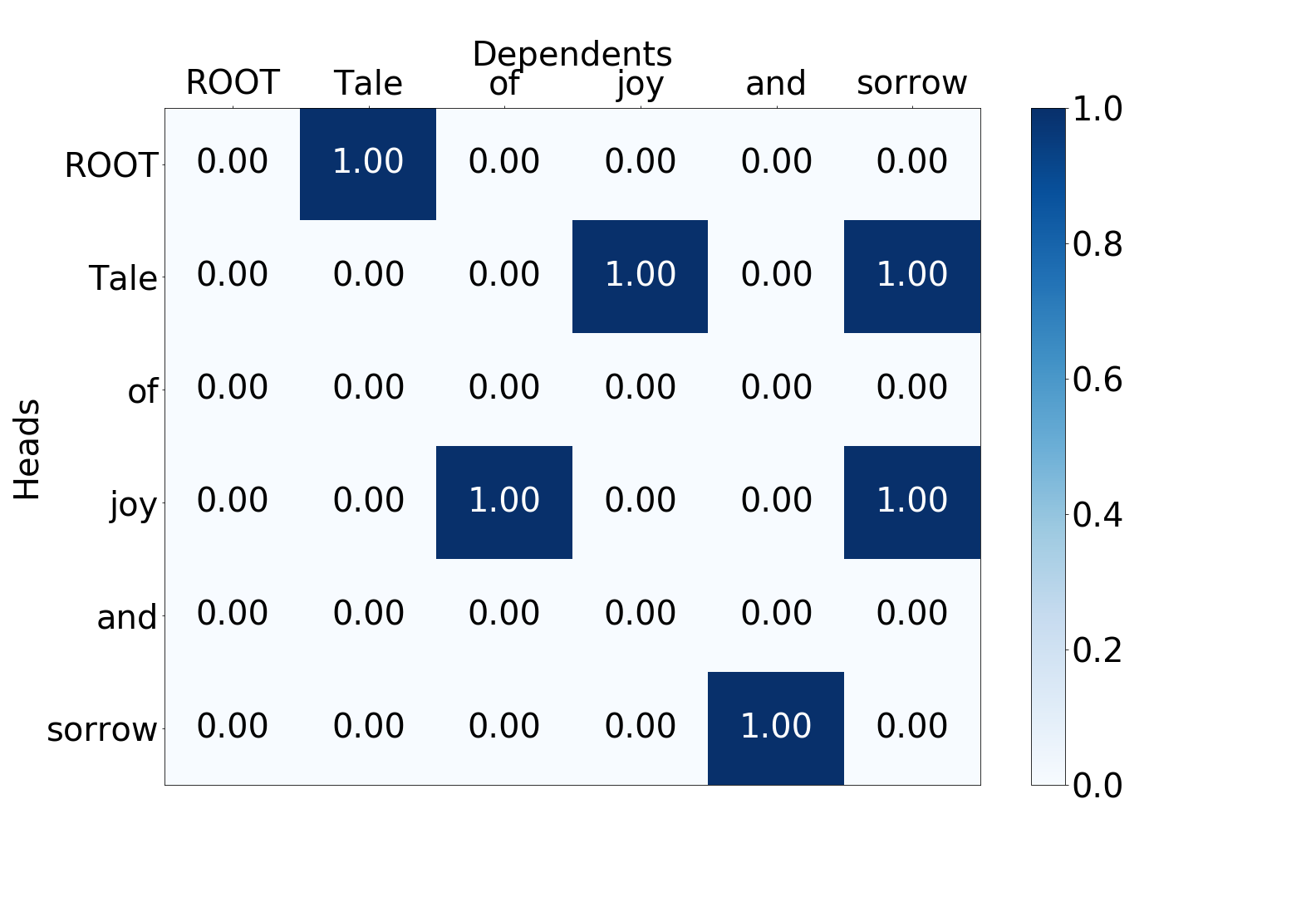}
   \caption{Edge-existence probabilities.
   }
   \label{fig:semanticb}
   \end{subfigure}
   \caption{
   The enhanced UD graph and edge-existence probabilities of the semantic parser trained on \texttt{en\_ewt} for the phrase \textit{Tale of joy and sorrow}.
    }
    \label{fig:semantic}
\end{figure*}

In order for the semantic dependency parser to be able to model relationships where a word may have multiple heads,
we create an adjacency matrix where the $ij^{th}$ entry in the matrix indicates whether an edge exists between tokens $i$ and $j$ with label type $k$.
We also append the dummy \textit{root} token to the adjacency matrix so that an edge can be
predicted from the main predicate of the sentence to the dummy \textit{root} token.

Figure~\ref{fig:semantica} shows the enhanced UD graph for the phrase, \emph{Tale of joy and sorrow}.
In the enhanced representation,
each conjunct in the conjoined noun phrase is attached to the governor of the modifier phrase,
e.g. there is an additional \textit{nmod} relation marked in blue between the noun \textit{Tale} and the second conjunct \textit{sorrow}.
Note that the lemma of the \textit{case} and \textit{cc} dependents
are appended to the enhanced dependency labels of their heads.
The corresponding edge-existence probabilities of the semantic parser trained on \texttt{en\_ewt} are shown in Figure~\ref{fig:semanticb} where the parser correctly predicts an edge from \emph{sorrow} to the first conjunct \emph{joy} as well as the head of the modifier phrase \emph{Tale}.

\subsubsection{Feature Representations}
In our experiments, each word $w_i$ in a sentence $ S=(w_0, w_1, \dots, w_N) $ is converted to its vector representation $\textbf{x}_i$.
We trained different variants of our semantic parser where
$\textbf{x}_i$ is the concatenation of different combinations of the below features:

\begin{itemize}
\item \textbf{BERT embedding}: 
The first word-piece embedding of the wordpiece-tokenised input word from BERT \citep{devlin-etal-2019-bert}
$ \mathbf{e}_i^{(BERT)} \in \mathbb{R}^{768}$
\item \textbf{character embedding}: 
A character embedding obtained by passing the $k$ characters $ch_1, \dots , ch_k$ of ${w_i}$ through a BiLSTM:
$ \text{BiLSTM}(ch_{1:k})$, $\mathbf{e}_i^{(ch)} \in \mathbb{R}^{64}$
\item \textbf{lemma embedding}:
The embedding of the word's lemma
$\mathbf{e}_i^{(le)} \in \mathbb{R}^{50}$
\item \textbf{UPOS embedding}:
The embedding of the word's universal POS tag
$\mathbf{e}_i^{(u)} \in \mathbb{R}^{50}$
\item \textbf{XPOS embedding}:
The embedding of the word's language-specific POS tag
$\mathbf{e}_i^{(x)} \in \mathbb{R}^{50}$
\item \textbf{morphological feature embedding}:
The embedding of the word's morphological features
$\mathbf{e}_i^{(f)} \in \mathbb{R}^{50}$
\item \textbf{head-information embedding}:
An embedding representing the word's head information from the basic tree
$\mathbf{e}_i^{(h)} \in \mathbb{R}^{50}$
\item \textbf{dependency label embedding}:
The embedding of the word's dependency label from the basic tree
$\mathbf{e}_i^{(label)} \in \mathbb{R}^{50}$
\end{itemize}

All model variants use the lexical information of the first BERT word-piece embedding and the character embedding, where $;$ represents vector concatenation:
\begin{equation}
    \label{eq:sdp1}
    \mathbf{e}_i^{(l)} = [\mathbf{e}_i^{(BERT)} ; \mathbf{e}_i^{(ch)}]
\end{equation}
The subsequent variation comes from the other types of features used where we experimented with the below feature settings:
\begin{equation}
    \label{eq:sdp2}
    \mathbf{x}_i = [\mathbf{e}_i^{(l)} ; \mathbf{e}_i^{(u)}]
\end{equation}
\begin{equation}
    \label{eq:sdp3}
    \mathbf{x}_i = [\mathbf{e}_i^{(l)} ; \mathbf{e}_i^{(le)} ; \mathbf{e}_i^{(u)} ; \mathbf{e}_i^{(f)}]
\end{equation}
\begin{equation}
    \label{eq:sdp4}
    \mathbf{x}_i = [\mathbf{e}_i^{(l)} ; \mathbf{e}_i^{(le)} ; \mathbf{e}_i^{(u)} ; \mathbf{e}_i^{(x)} ; \mathbf{e}_i^{(f)}]
\end{equation}
\begin{equation}
    \label{eq:sdp5}
    \mathbf{x}_i = [\mathbf{e}_i^{(l)} ; \mathbf{e}_i^{(le)} ; \mathbf{e}_i^{(u)} ; \mathbf{e}_i^{(f)} ; \mathbf{e}_i^{(b)}]
\end{equation}
\begin{equation}
    \label{eq:sdp6}
    \mathbf{x}_i = [\mathbf{e}_i^{(l)} ; \mathbf{e}_i^{(le)} ; \mathbf{e}_i^{(u)} ; \mathbf{e}_i^{(x)} ; \mathbf{e}_i^{(f)} ; \mathbf{e}_i^{(b)}]
\end{equation}

For the morphological features, there may be multiple morphological tags $m_1, \dots , m_t$ for a particular word $w_i$.
Thus, we split the full label into separate features \citep{hall-etal-2007-single} and embed each morphological property separately.
We then sum the individual embedded representations
together and divide by the number of active properties:

\begin{equation}
    \label{eq:sdp7}
    \mathbf{e}_i^{(f)} =  \text{mean}(\mathbf{e}^{(m_{1:t})})
\end{equation}

We follow the same process for the head-information embedding $\mathbf{e}_i^{(h)}$.
Rather than encoding the head as an integer value, we obtain a \emph{direction} value and a \emph{distance} value:
for each head-dependent pair $(i, j)$, we subtract the indices of $i, j$ giving the distance value.
If the value is negative it means the head is to the left or if it is positive, to the right.
We then take the absolute distance value and define ranges:
\textit{short} (1-4),
\textit{medium} (5-9),
\textit{far} (10-14)
and \textit{long-range} ($>$15).
The qualitative direction (\textit{left} or \textit{right}) and distance labels are embedded in the same way as morphological features, e.g. embedded as separate components, summed together and then divided by the number of features (which in this case is always two):
\begin{equation}
    \label{eq:sdp8}
    \mathbf{e}_i^{(h)} =  \text{mean}(\mathbf{e}^{(h_{1:t})})
\end{equation}
To encode the basic tree, we then concatenate the head representation and the dependency label embedding:
\begin{equation}
    \label{eq:sdp9}
    \mathbf{e}_i^{(b)} =  [\mathbf{e}_i^{(h)} ; \mathbf{e}_i^{(label)}]
\end{equation}
It is worth mentioning that more sophisticated approaches for modelling head distance and direction exist for basic dependency parsing \citep{qi-etal-2018-universal} but we leave using this approach for enhanced dependency parsing as future work.

\subsubsection{Training Details}
Our semantic parser predicts edges in a
greedy fashion based on local decisions,
\ie we did not make use of any maximum spanning tree algorithm 
or enforce any global constraints.
One property of enhanced dependency graphs is that the graph may contain cycles, therefore, we did not remove any cycles from the graph but observed that this sometimes causes fragments in the graph which are not reachable from the notional ROOT.
For graphs with unreachable nodes, we applied our post-processor to attach these (Section~\ref{sec:connect}).

We found that this architecture can be easily applied to enhanced dependency parsing given its similar nature to semantic dependency parsing.
One caveat is that in enhanced dependency parsing, the label set can be quite large as modifier lemma and case information can be appended to the dependency label which
results in very high memory requirements for certain languages such as Arabic.
Additionally, 
modelling all enhanced labels in this fashion means that the parser is limited in its ability to predict
labels for rare modifiers.
An examination of the semantic parser output on the \texttt{en\_ewt} development set shows that, although the parser often predicts the correct label, it can sometimes predict the wrong label containing a frequent modifier which is not in the sentence, e.g. \textit{advcl:if} instead of 
\textit{advcl:as}.

\begin{table}[t!]
\begin{center}
\begin{tabular}{ll}
\toprule
\multicolumn{2}{c}{\bf Semantic Parser Details } \\
\bf Parameter        & \bf Value \\
\midrule
Char-BiLSTM layers & 2 \\
BiLSTM layers      & 3 \\
BiLSTM size        & 400 \\
Char-BiLSTM size        & 64 \\
Arc MLP size     & 500 \\
Label MLP size & 100 \\
Dropout LSTMs      & 0.33 \\
Dropout MLP        & 0.33 \\
Dropout embeddings & 0.33 \\
Nonlinear act. (MLP) & ELU \\
Edge prediction threshold & 0.5 \\
BERT word-piece embedding       & 768 \\
Char embedding       & 64 \\
Tag embedding (all tags)      & 50 \\
Optimizer          & Adam\\
Learning rate      & 0.001\\
beta1           & 0.9\\
beta2              & 0.9\\
Num. epochs              & 75 \\
Patience              & 10 \\
Batch size              & 16 \\
\bottomrule
\end{tabular}
\end{center}
\caption{\label{table:hyper-params} Chosen hyperparameters for our semantic parser.
For the tag embedding, we use the same size embedding for all features (lemma, POS, morphological features, head-information and label embeddings) and concatenate them.}
\end{table}

Our semantic parser is built upon the implementation in AllenNLP \citep{gardner-etal-2018-allennlp}.
Due to time constraints,
we trained our semantic parsing models on the gold training data released by the organisers
as opposed to
creating jack-knifed silver data. 
Hyperparameters are similar to those in \newcite{dozat-manning-2017-deep} as we found the larger network size of \newcite{dozat-manning-2018-simpler} to be too restrictive for certain languages with high memory demands.
Full hyperparameters of the semantic parser are given in Table~\ref{table:hyper-params}.
We trained 
for 75 epochs with early-stopping if the development score did not improve after 10 epochs.

\paragraph{Memory Considerations}
We trained our semantic parsing models on two GPUs:
the first was an NVIDIA RTX 2080 Ti with 12GB of VRAM
where we had to remove very long sentences ($<.03\%$ of sentences overall) from the treebanks: \texttt{cs\_cac}, \texttt{cs\_pdt}, \texttt{it\_isdt}, \texttt{ru\_syntagrus} and \texttt{sv\_talbanken} in order to fit a batch into memory.
We were also given access to an NVIDIA V100 GPU with 32GB of VRAM which enabled us to process all treebanks except for \texttt{ar\_padt} without removing long sentences.
For \texttt{ar\_padt}, after removing the longest 75 sentences, the model still required 29GB of VRAM.

\subsubsection{BERT Models}

For the BERT models,
in early development runs we compared multilingual BERT (mBERT) with a language-specific BERT model if there was one available in HuggingFace's \citep{wolf-etal-2019-huggingface}
models repository.\footnote{\url{https://huggingface.co/models}}
We used a language-specific BERT model for
\texttt{ar} \citep{safaya-etal-2020-kuisail},
\texttt{bg+cs} \citep{arkhipov-etal-2019-tuning},
\texttt{en} \citep{devlin-etal-2019-bert},
\texttt{fi} \citep{virtanen-etal-2019-multilingual},
\texttt{it}\footnote{\url{https://github.com/dbmdz/berts}},
\texttt{nl} \citep{vries-etal-2019-bertje},
\texttt{pl}\footnote{\url{https://github.com/kldarek/polbert}},
\texttt{ru} \citep{kuratov-arkhipov-2019-adaptation} 
and \texttt{sv}\footnote{\url{https://github.com/Kungbib/swedish-bert-models}}
and for the rest of the languages we used mBERT \citep{devlin-etal-2019-bert}.
We found that the language-specific variant was always better than mBERT except for \texttt{pl\_lfg}.
For \texttt{fr\_sequoia}, we tried using the CamemBERT model \citep{martin-etal-2020-camembert}.
As this model uses RoBERTA \citep{liu-etal-2019-roberta}
as opposed to BERT, we installed AllenNLP from the master repository which uses HuggingFace's \texttt{AutoTokenizer} module which supports many BERT-like models.
We noticed a trend of lower results when using the master branch for some languages
but training was also more stable for certain treebanks where we had previously encountered a \texttt{nan} in the loss.\footnote{We incurred a \texttt{nan} loss for \texttt{cs\_cac}, \texttt{cs\_pdt}, \texttt{it\_isdt} and \texttt{ru\_syntagrus} using the AllenNLP stable branch 0.9.0 and used the best model from the available epochs.}
Consequently, we include models from the stable release and the bleeding-edge master branch in our development pipeline.


\subsection{Connecting the Graph}
\label{sec:connect}

We had no solution ready to connect fragmented graphs
produced by our semantic parser\footnote{Between
    90.18\% (Lithuanian) and 99.51\% (Russian) 
    of test sentences in our official submission are not affected,
    \ie all nodes are reachable from a root node.
    This observation excludes Estonian, for which we submitted
    predictions using our heuristic system.
}
on the system submission day
and resorted to using the
``connect-to-root'' option of the \texttt{quick-fix} tool provided 
by the shared task organisers, who warned that it
had not been thoroughly tested.

After the system submission deadline, we investigated the fragmentation
issue.
The task is to make all nodes reachable from 
the notional ROOT\footnote{UD distinguished between
    the notional ROOT (ID 0) and root nodes. The latter are any nodes
    that have `0' as a head.
}, where reachability is directional.
Adding more edges than necessary harms precision and thus F1-score.
We found that the \texttt{quick-fix} tool with the
``connect-to-root'' option adds edges to every unreachable node.
We also noticed a bug in the implementation where certain reachable nodes were
being reported as unreachable.

We then
implemented an improved tool to connect fragmented graphs trying to minimise
the number of edges added to the graph.
We repeatedly check for each unreachable node how many unreachable nodes
can be reached from it.
Among the nodes that maximise this number we
pick the first node in surface order and make it a child of
the notional ROOT, \ie it becomes an additional root node.
This is a rather naive approach which does not try to connect fragments in a sensible manner but, rather, mimics the behaviour of the ``connect-to-root'' option.
Future work could try to show whether our above algorithm adds the minimal
number of edges necessary to connect the graph or if a lower optimum exists.


\section{Results}
\label{sec:results}


\begin{table}
\centering
\begin{tabular}{l|rr}
\toprule
 & \multicolumn{2}{c}{\textbf{ELAS F1}} \\
\textbf{Treebank} & \textbf{sem-frag}
 & \textbf{heuristic}\\
\midrule
ar\_padt & \textbf{70.99} & 59.74 \\
bg\_btb & \textbf{88.09} & 86.19 \\
\midrule
cs\_cac & \textbf{86.51} & 74.41 \\
cs\_fictree & \textbf{83.23} & 77.37  \\
cs\_pdt & \textbf{79.58} & 	71.19 \\
\midrule
en\_ewt & \textbf{84.71} & 	82.86 \\
et\_edt & 62.74 & \textbf{69.35} \\
fi\_tdt & \textbf{83.64} & 71.84 \\ 
fr\_sequoia & \textbf{88.65} &  87.53 \\
it\_isdt & \textbf{90.13} & 88.28 \\
lt\_alksnis & \textbf{73.63} & 57.84 \\
lv\_lvtb & \textbf{81.82} & 71.29 \\
\midrule
nl\_alpino & \textbf{89.93} & 89.00 \\
nl\_lassysmall & 79.00 & \textbf{81.24} \\
\midrule
pl\_lfg & \textbf{94.12} & 93.84 \\
pl\_pdb & \textbf{82.25} & 78.27 \\
\midrule
ru\_syntagrus & \textbf{88.48} & 80.03 \\
sk\_snk  & \textbf{81.30} & 75.98 \\
sv\_talbanken & \textbf{84.54} & 81.32 \\
ta\_ttb & \textbf{55.68} & 43.94 \\
uk\_iu & \textbf{82.41} & 76.88 \\
\bottomrule
\end{tabular}
\caption{Development set ELAS F1 score 
        for the best semantic parser evaluated without connecting
            fragmented graphs (sem-frag)
        and
        for the best combination of heuristic rules
            (heuristic)
}
\label{devresults:decision_custom}
\end{table}


Table~\ref{devresults:decision_custom} compares the semantic parser against the heuristic approach on the ELAS F1 metric.
The evaluation script was run without connecting fragmented graphs and format validation.
For all but two treebanks, the semantic parser performs better than the
best
heuristic approach.
For some languages, the difference in performance is large.
For \texttt{et\_ewt}, which does not have a development set,
we suspect that we overfitted our semantic parser on the
\texttt{et\_ewt} training data
by allowing it to train for 75 epochs.



\begin{table}
\centering
\begin{tabular}{l|rrr}
\toprule
 & \multicolumn{3}{c}{\textbf{ELAS F1}} \\
\textbf{Treebank} & \textbf{subm}
 & \textbf{frag fix} & \textbf{re-run}\\
\midrule
Arabic-PADT         &  57.19  &  70.08  &  \bf 70.40  \\
Bulgarian-BTB       &  77.29  &  89.58  &  \bf 89.60  \\
Czech-FicTree       &  70.04  &  80.77  &  \bf 81.63  \\
Czech-CAC           &  71.72  &  86.00  &  \bf 86.38  \\
Czech-PDT           &  65.94  &  79.03  &  \bf 79.84  \\
Czech-PUD           &  64.34  &  77.37  &  \bf 78.08  \\
Dutch-Alpino        &  71.44  &  87.61  &  \bf 87.77  \\
Dutch-L.Small       &  64.03  &  77.39  &  \bf 77.24  \\
English-EWT         &  70.61  &  \bf 83.56  & \bf 83.56  \\
English-PUD         &  70.25  &  86.88  & \bf 87.03  \\
Estonian-EDT        &  62.29  &  68.20  &  \bf 68.37  \\
Estonian-EWT        &  55.70  &  \bf 61.19  &  60.67  \\
Finnish-TDT         &  73.02  &  \bf 84.36  &  84.33  \\
Finnish-PUD         &  71.58  & \bf 84.62  & \bf 84.62  \\
French-Sequoia      &  77.44  &  87.58  & \bf 88.60  \\
French-FQB          &  74.30  &  82.68  & \bf 83.26  \\
Italian-ISDT        &  71.98  &  \bf 90.24  &  90.23  \\
Latvian-LVTB        &  72.41  &  81.81  &  \bf 82.40  \\
Lithuanian-AL.      &  58.36  &  68.76  &  \bf 68.84  \\
Polish-LFG          &  61.23  &  \bf 70.89  &  70.71  \\
Polish-PDB          &  67.68  &  80.93  &  \bf 82.43  \\
Polish-PUD          &  65.64  &  79.77  & \bf 80.79  \\
Russian-SynT.       &  75.27  &  89.21  & \bf 89.47  \\
Slovak-SNK          &  68.43  &  81.63  &  \bf 81.97  \\
Swedish-Talb.       &  71.86  &  86.78  & \bf 86.72  \\
Swedish-PUD         &  64.70  &  79.35  & \bf 79.37  \\
Tamil-TTB           &  48.47  &  \bf 57.28  &  57.10  \\
Ukrainian-IU        &  66.43  &  79.81  & \bf 82.92  \\
\midrule
Average             &  67.49  &  79.76  & \bf 80.15  \\
\bottomrule
\end{tabular}
\caption{Test set results:
    subm = submitted,
    frag fix = using our own fragment connector and quick-fix.pl without connect-to-root,
    re-run = a re-run with bug fixes, no new models but new model selection
}
\label{testresults_custom}
\end{table}


Table~\ref{testresults_custom} shows test set ELAS obtained on the shared task
submission site for
\textit{(a)} our submission fully relying on the organiser's
             \texttt{quick-fix} tool to fix issues in the output of
             our system,
\textit{(b)} the same predictions post-processed by our own
             fragment connector that aims to minimise the
             number of root edges added, and
\textit{(c)} a re-run of our pipeline using the same models
             for system components as before but with all
             bugs fixed during development applied to all
             predictions and new decisions which models
             to apply to the test sets.
While the \texttt{quick-fix} tool enabled us to make a valid submission
in time, its
approach of adding edges from the root node to
all unreachable tokens
has a strong negative impact on 
precision, \eg 62.26 ELAS precision on the Czech CAC development set
\vs 87.37 without post-processing.
Our own post-competition fix avoids this
and would have brought us to the top half of the competition.


\section{Conclusion}
\label{sec:conclusion}

In this system submission,
we use a graph-based semantic parser to parse enhanced dependencies and compare to a baseline in which we create enhanced graphs from the basic tree using a very limited set of heuristics.
Avenues for future work include:
    
    \paragraph{Post-processing} Predict the head and label for edges connecting fragments (as opposed to a dummy ``0:root'' edge) where this information could come from new edges available from lowering the score threshold or from the basic tree.
    \paragraph{Label Prediction} The semantic parser  performs competitively despite treating enhanced dependency labels containing lemmas and case information 
    as atomic units. However, a more sophisticated approach should still be tried.

    \paragraph{Multi-treebank Parsing} When randomising the proxy treebank for multi-treebank models,
          use a different randomisation for each ensemble member.
    Predict the best proxy treebank for each test sentence or paragraph \cite{wagner-etal-2020-treebank}.
    
    \paragraph{Elided Tokens} Our semantic parser handles elided tokens by appending the elided token to the adjacency matrix and offsetting the head indices.
    While we used this approach during training on gold data, we did not predict elided tokens 
    and we wish to explore methods for doing so.



\section*{Acknowledgments}

This  research  is  supported  by  Science  Foundation Ireland
through the ADAPT Centre for Digital Content Technology, which is
funded under the SFI Research Centres Programme (Grant 13/RC/2106)
and is co-funded under the European Regional Development Fund.
We thank the reviewers for their
insightful, detailed feedback.
We acknowledge Dell for the use of an NVIDIA V100 GPU as part of the Dell AI Ready Bundle with Nvidia.


\bibliographystyle{acl_natbib}
\bibliography{main}

\end{document}